\documentclass[conference]{IEEEtran}
\IEEEoverridecommandlockouts

\usepackage{cite}
\usepackage{amsmath,amssymb,amsfonts}
\usepackage{algorithmic}
\usepackage{graphicx}
\usepackage{textcomp}
\usepackage{xcolor}

\usepackage[citecolor=blue,colorlinks=true,urlcolor=blue,linkcolor=blue,bookmarks=false]{hyperref}
\usepackage[colorinlistoftodos]{todonotes}
\usepackage{url}
\usepackage{algorithm}
\usepackage{algorithmic}
\usepackage{lipsum}
\usepackage[keeplastbox]{flushend}

\usepackage{booktabs}

\usepackage{mathtools}
\DeclarePairedDelimiter{\norm}{\lVert}{\rVert}

\newtheorem{definition}{Definition}
\usepackage{amsfonts}
\usepackage{amsmath}
\usepackage{subfig}
\title{\LARGE \bf
Adversarial Attacks on Deep Neural Networks\\ for Time Series Classification
}

\author{\IEEEauthorblockN{Hassan Ismail Fawaz,
Germain Forestier,
Jonathan Weber,
Lhassane Idoumghar and 
Pierre-Alain Muller\smallskip }
\IEEEauthorblockA{IRIMAS,
Universit\'e Haute-Alsace, Mulhouse, France \smallskip}
\IEEEauthorblockN{\textit{Email}: \{first-name.last-name@uha.fr\}}
}

\begin{document}
\bstctlcite{IEEEexample:BSTcontrol}
\maketitle
\thispagestyle{empty}
\pagestyle{empty}

\begin{abstract} \small\baselineskip=9pt 
Time Series Classification (TSC) problems are encountered in many real life data mining tasks ranging from medicine and security to human activity recognition and food safety.
With the recent success of deep neural networks in various domains such as computer vision and natural language processing, researchers started adopting these techniques for solving time series data mining problems.   
However, to the best of our knowledge, no previous work has considered the vulnerability of deep learning models to adversarial time series examples, which could potentially make them unreliable in situations where the decision taken by the classifier is crucial such as in medicine and security.
For computer vision problems, such attacks have been shown to be very easy to perform by altering the image and adding an imperceptible amount of noise to trick the network into wrongly classifying the input image.
Following this line of work, we propose to leverage existing adversarial attack mechanisms to add a special noise to the input time series in order to decrease the network's confidence when classifying instances at test time. 
Our results reveal that current state-of-the-art deep learning time series classifiers are vulnerable to adversarial attacks which can have major consequences in multiple domains such as food safety and quality assurance. 
\end{abstract}

\section{Introduction}
Time Series Classification (TSC) problems are encountered in various real world data mining tasks ranging from health care~\cite{abdelfattah2018augmenting,ma2018health,IsmailFawaz2018evaluating} and security~\cite{tan2017indexing,tobiyama2016malware} to food safety~\cite{briandet1996discrimination,nawrocka2013determination} and power consumption monitoring~\cite{owen2012powering,zheng2018wide}.
As deep learning models have revolutionized many machine learning fields such as computer vision~\cite{krizhevsky2012imagenet} and natural language processing~\cite{yang2018investigating,wang2018hierarchical}, researchers recently started to adopt these models for TSC tasks~\cite{ismailfawaz2018deep}.

Following the advent of deep learning, researchers started to study the vulnerability of deep networks to adversarial attacks~\cite{yuan2017adversarial}.
In the context of image recognition, an adversarial attack consists in modifying an original image so that the changes are almost undetectable by a human~\cite{yuan2017adversarial}.
The modified image is called an adversarial image, which will be misclassified by the neural network, while the original one is correctly classified. 
One of the most famous real-life attacks consists in altering a traffic sign image so that it is misinterpreted by an autonomous vehicle~\cite{eykholt2018robust}.
Another application is the alteration of illegal content to make it undetectable by automatic moderation algorithms~\cite{yuan2017adversarial}.
The most common attacks are gradient-based methods, where the attacker modifies the image in the direction of the gradient of the loss function with respect to the input image thus increasing the misclassification rate~\cite{goodfellow2015explaining,kurakin2017adversarial,yuan2017adversarial}. 

\begin{figure}
\centering
\includegraphics[width=.95\linewidth]{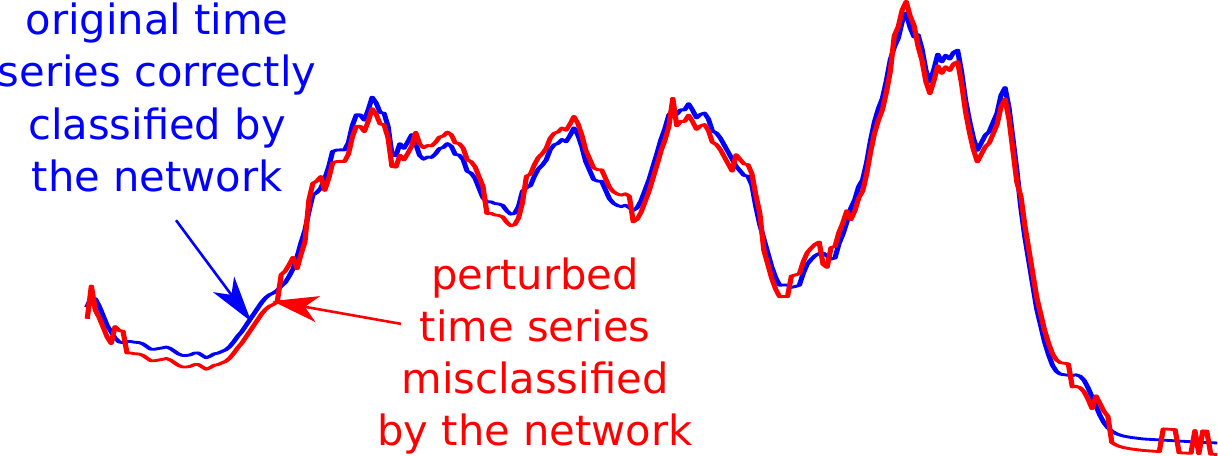}
\caption{Example of a perturbed time series that is misclassified by a deep network after applying a small perturbation (time series from the Coffee dataset~\cite{chen2015ucr} containing spectrographs of coffee beans).}
\label{fig-cd-diagram-mts-ucr}
\end{figure}

While these approaches have been intensely studied in the context of image recognition (e.g. NIPS competition on Adversarial Vision Challenge), adversarial attacks haven not been thoroughly explored for TSC.
This is surprising as deep learning models are getting more and more popular to classify time series~\cite{IsmailFawaz2019deep,wang2017time,ma2018health,zheng2018wide,IsmailFawaz2018data,IsmailFawaz2018transfer}.
Furthermore, potential adversarial attacks are present in many applications where the use of time series data is crucial.
For example, Figure~\ref{fig-cd-diagram-mts-ucr} shows an original and perturbed time series of coffee beans spectrographs. 
While a deep neural network correctly classifies the original time series as Robusta beans, adding small perturbations makes it classify it as Arabica.
Therefore, since Arabica beans are more valuable than Robusta beans, this attack could be used to deceive food control tests and eventually the consumers.

In this paper, we present, transfer and adapt adversarial attacks that have been shown to work well on images to time series data. 
We also present an experimental study using the 85 datasets of the UCR archive~\cite{chen2015ucr} which reveals that neural networks are prone to adversarial attacks.
We highlight specific real-life use cases to stress the importance of such attacks in real-life settings, namely food quality and safety, vehicle sensors and electricity consumption.
Our findings show that deep networks for time series data are vulnerable to adversarial attacks like their computer vision counterparts.
Therefore, this paper sheds the light on the need to protect against such attacks, especially when deep learning is used for sensitive TSC applications. 
We also show that adversarial time series learned using one network architecture can be transferred to different architectures.
Finally, we discuss some mechanisms to prevent these attacks while making the models more robust to adversarial examples.

\noindent To summarize, the main contributions of this paper are:
\begin{itemize}
    \item The definition and formalization of adversarial attacks for TSC tasks.
    \item The transfer and adaptation of image-based attacks to time series data.
    \item An empirical study of these methods on the UCR archive datasets.
    \item A set of use-cases that highlight the importance of such attacks in real-life scenarios.
    \item An open-source framework to generate adversarial time series.
    \item A set of adversarial time series for each dataset in the UCR archive.
    \item A list of open issues to be considered in future research on this topic.
\end{itemize}

\section{Background}
In this section, we start with the necessary definitions for ease of understanding. 
We then follow by an overview of critical applications based on deep learning approaches for TSC where adversarial attacks could have serious and dangerous consequences.
Finally, we present a brief survey of the current state-of-the-art methods for adversarial attacks which have been mainly proposed and validated on image datasets~\cite{yuan2017adversarial}.  

\begin{definition}
A time series $X=[x_1,x_2, \dots, x_T]$ is an ordered set of real values. 
The length of $X$ is equal to the number of real values $T$. 
\end{definition}

\begin{definition}
$D=\{(X_1,Y_1), \dots, (X_N,Y_N)\}$ is a dataset of pairs $(X_i,Y_i)$ where $X_i$ is a time series with $Y_i$ as its corresponding one-hot label vector.
\end{definition}

\begin{definition}
Time Series Classification (TSC) task consists of training a classifier on $D$ in order to map from the space of possible inputs to a probability distribution over the class variable values (labels).
\end{definition}

\begin{definition}
$f(\cdot) \in F: \mathbb{R}^T \rightarrow \hat{Y} $ represents a deep learning model for TSC.
\end{definition}

\begin{definition}
$J_f(\cdot , \cdot)$ denotes the loss function (e.g.\ cross-entropy) of the model $f$.
\end{definition}

\begin{definition}\label{def-adv-ex}
$X^{'}$ denotes the adversarial example, a perturbed version of $X$ (the original instance) such that $\hat{Y}\ne \hat{Y}^{'}$ and $\norm{X-X^{'}}_p\leq \epsilon$.
\end{definition}

\subsection{Deep learning for time series classification}
Since AlexNet~\cite{krizhevsky2012imagenet} won the ImageNet competition in 2012, deep learning has seen a lot of successful applications in various domains such as reaching human level performance in image recognition problems~\cite{ismailfawaz2018deep} as well as different natural language processing tasks~\cite{yang2018investigating}. 
Motivated by this success, researchers and data mining practitioners started adopting these deep machine learning models in various real life TSC problems~\cite{ma2018health,zheng2018wide,tobiyama2016malware,cabral2018an,wang2017time,ismailfawaz2018deep}.
In fact, deep Convolutional Neural Networks (CNNs) were shown to achieve state-of-the-art performance for TSC when evaluated over the UCR archive benchmark.
Specifically, by sliding one dimensional filters over the input time series, the network is able to learn discriminative, non-linear and time invariant features useful for classification.
For a more comprehensive description on how deep CNNs are being adapted for one dimensional temporal data, we refer the interested reader to our recent empirical study of deep learning models for TSC~\cite{ismailfawaz2018deep}.

In this paper, we focus on the application of deep neural networks in crucial and sensitive decision making systems, thus motivating the investigation of neural network's vulnerabilities to adversarial examples.  
In~\cite{ma2018health}, CNNs were used to mine temporal electronic health data for risk prediction and disease sub-typing.  
In situations where algorithms are taking the decision for reimbursement of medical treatment, tampering medical records in an imperceptible manner could eventually lead to fraud.    
Apart from the health care industry, deep CNNs are also being used when monitoring power consumption from houses or factories. 
For example in~\cite{zheng2018wide}, time series data from smart grids were analyzed for electricity theft detection, where in such use cases perturbed data can help thieves to avoid being detected.
Other crucial decision making systems such as malware detection in smart-phones, leverage temporal data in order to classify if an Android application is malicious or not~\cite{tobiyama2016malware}. 
Using adversarial attacks, a hacker might generate synthetic data from his/her application allowing it to bypass the security systems and get it installed on the end user's smart-phone.
Finally, when deep neural networks are deployed for road anomaly detection~\cite{cabral2018an}, perturbing the data recorded by sensors placed on the road could help the entities responsible for such life threatening anomalies, to avoid being captured. 
We should note that due to space limitations, this list of potential attacks is not exhaustive.

\begin{figure*}[th!]
\centering
\includegraphics[width=\linewidth]{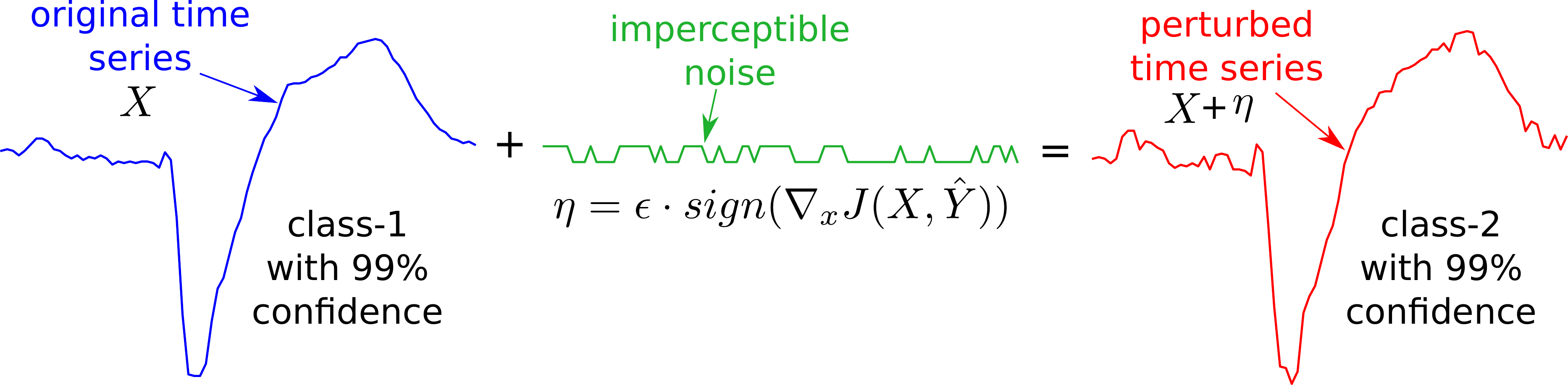}
\caption{Example of perturbing the classification of an input time series from the TwoLeadECG dataset by adding an imperceptible noise computed using the Fast Gradient Sign Method (FGSM).}
\label{fig-pert-example}
\end{figure*}

\subsection{Adversarial attacks}
Szegedy~\emph{et~al.}~\cite{szegedy2014intriguing} were the first to introduce adversarial examples against deep neural networks for image recognition tasks in 2014. 
Following these intriguing findings, a huge amount of research has been dedicated to generating, understanding and preventing adversarial attacks on deep neural networks~\cite{goodfellow2015explaining,kurakin2017adversarial,eykholt2018robust}.   

Most of these methods have been proposed for image recognition tasks~\cite{yuan2017adversarial}. 
For example, in~\cite{goodfellow2015explaining}, a fast gradient-based attack was developed as an alternative to expensive optimization techniques~\cite{szegedy2014intriguing}, where the authors explained the presence of such adversarial examples with the hypothesis of linearity for deep learning models.
This kind of attack was also extended by a more costly iterative procedure~\cite{kurakin2017adversarial}, where the authors showed for the first time that even printed adversarial images (\emph{i.e.} perceived by a camera) are able to fool a pre-trained network.   
More recently, it has been shown that perturbing stop signs can trick autonomous vehicles into misclassifying it as a speed limit sign~\cite{eykholt2018robust}. 

Other fields such as Natural Language Processing have also been investigated to create adversarial attacks such as adding distracting phrases at the end of a paragraph in order to show that deep learning-based reading comprehension systems were not able to distinguish subtle differences in text~\cite{robin2017adversarial}.  
For a review on the different adversarial attacks for deep learning systems, we refer the interested readers to a recent survey in~\cite{yuan2017adversarial}.

For general TSC tasks, it is surprising how adversarial attack approaches have been ignored by the community.
The only previous work mentioning attacks for TSC is~\cite{oregi2018adversarial}. 
By adapting a soft $K$ Nearest Neighbors (KNN) coupled with Dynamic Time Warping (DTW), the authors showed that adversarial examples could fool the proposed nearest neighbors classifier on a single simulated dataset (synthetic\_control from the UCR archive~\cite{chen2015ucr}). 
However, the fact that the KNN classifier is no longer considered as the state-of-the-art classifier for time series data~\cite{bagnall2017the}, we believe that it is important to investigate the generation of adversarial time series examples that deteriorate the accuracy of state-of-the-art classifiers such a Residual Network (ResNet)~\cite{ismailfawaz2018deep,wang2017time} and to validate it on the whole 85 datasets in the UCR archive.
Finally, we formally define an adversarial attack on deep neural networks for TSC.
\begin{definition}
Given a trained deep learning model $f$ and an original input time series $X$, generating an adversarial instance $X^{'}$ can be described as a box-constrained optimization problem.
\begin{multline}\label{eq-optim}
    \min_{X^{\theta}} \norm{X^{'}-X}~s.t.\\
     f(X^{'})=\hat{Y^{'}},~f(X)=\hat{Y}~and~\hat{Y}\ne\hat{Y^{'}}
\end{multline}
\end{definition}
Let $\eta=X-X^{'}$ be the perturbation added to $X$, which corresponds to a very low amplitude signal. 
Figure~\ref{fig-pert-example} illustrates this process where the green time series corresponds to the added perturbation $\eta$.
The optimization problem in (\ref{eq-optim}) minimizes the perturbation while misclassifying the input time series.

\section{Adversarial attacks for time series}
In this section, we describe the ResNet architecture and present two attack methods that we then use to generate adversarial time series examples for the ResNet model. 

\begin{figure*}[t]
\centering
\includegraphics[width=\linewidth]{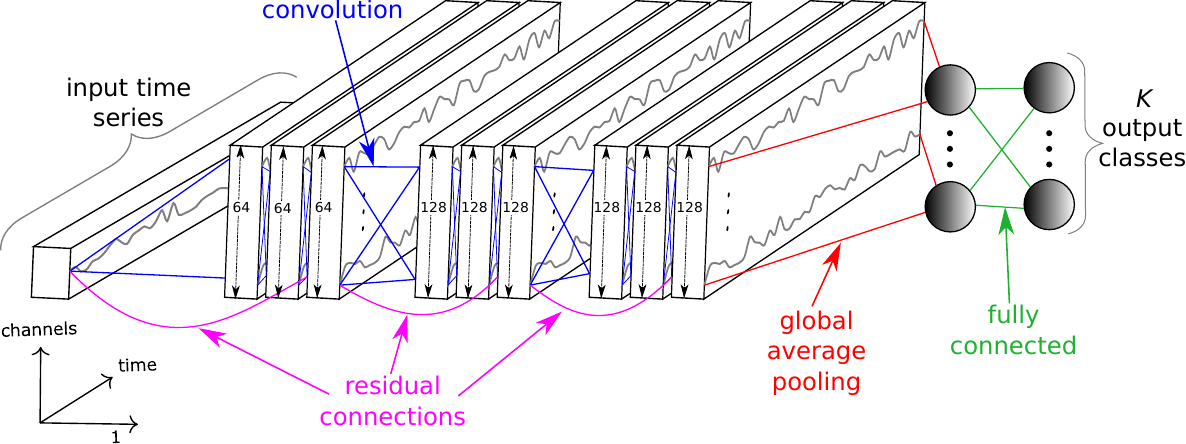}
\caption{The deep Residual Network (ResNet) architecture for Time Series Classification (TSC).}
\label{fig-resnet-archi}
\end{figure*}


\subsection{Residual Network} 
ResNet was originally proposed for TSC in~\cite{wang2017time}, where it was validated on 44 datasets from the UCR archive~\cite{chen2015ucr}. 
In a recent empirical evaluation~\cite{ismailfawaz2018deep} on the 85 datasets from the UCR archive, we identified that ResNet achieved state-of-the-art performance for TSC, with results that are not significantly different than the Collective Of Transformation-based Ensembles, the current state-of-the-art classifier, which is an ensemble of 35 classifiers~\cite{bagnall2017the}. 
Note that our adversarial attack methods are independent of the chosen network architecture, and that we chose ResNet for its robustness~\cite{ismailfawaz2018deep} as well as its use in many critical domains such as malware detection~\cite{cabral2018an}.
In addition, adversarial examples are known to be transferable across different neural network architectures which enables the synthetic time series to fool other deep learning models: a technique known as black-box attack~\cite{yuan2017adversarial}.

Figure~\ref{fig-resnet-archi} illustrates the adopted architecture in our adversarial attack experiments. 
The network's input is a time series of length $T$.
The output of the network is a probability distribution over the $K$ possible classes in the dataset. 
The first nine layers are convolutional layers with the Rectified Linear Unit (ReLU) as activation function~\cite{krizhevsky2012imagenet}.
Each convolutional layer is followed by a batch normalization operation~\cite{ioffe2015batch}.
In each convolutional block, the first, second and third convolutions are composed respectively of filters of length 8, 5 and 3.
The first, second and last convolutional blocks are comprised respectively of 64, 128 and 128 filters.
For all convolutions the stride is set to 1. 

Each convolutional layer takes as input a time series and applies a non-linearity to transform it into a multivariate time series whose dimensions are inferred from the number of filters in each layer. 
The tenth layer is composed of a Global Average Pooling (GAP) operation which takes the output
of the third convolutional block and averages the time series over the time dimension. 
This averaging operation reduces drastically the number of parameters in a deep model while enabling the use of a class activation map which allows an interpretation of the learned features~\cite{wang2017time}. 
The output of layer ten is then fed to a softmax classification layer whose number of neurons is equal to the number of classes $K$ in the dataset.
The network is trained for 1500 epochs using Adam~\cite{kingma2015adam} (with default hyperparameters) to minimize the objective cost function: cross-entropy.  

\subsection{Fast Gradient Sign Method}
FGSM was first proposed in~\cite{goodfellow2015explaining} to generate adversarial images that fooled the famous GoogLeNet model.
FGSM is considered  ``fast'' and replaces the expensive linear search method previously proposed in~\cite{szegedy2014intriguing}. 
The attack is based on a one step gradient update along the direction of the gradient's sign at each time stamp. 
The perturbation process  (illustrated in Figure~\ref{fig-pert-example}) can be expressed as:
\begin{equation}
\eta = \epsilon \cdot sign(\nabla_x J(X,\hat{Y}))  
\end{equation}
where $\epsilon$ denotes the magnitude of the perturbation (a hyperparameter). 
The adversarial time series $X^{'}$ can be easily generated with $X^{'}=X+\eta$. 
The gradient can be efficiently computed using back-propagation.

\subsection{Basic Iterative Method}
BIM extends FGSM by applying it multiple times with a small step size and clip the obtained time series elements after each step to ensure that they are in an $\epsilon$-neighborhood of the original time series~\cite{kurakin2017adversarial}. 
In fact, by adding smaller changes or perturbations in an iterative manner, the method is able to generate adversarial examples that are closer to the original samples and have a better chance of fooling the network. 
Algorithm~\ref{algo-bim} shows the different steps of this iterative attack which requires setting three hyperparameters: (1) the number of iterations $I$; (2) the amount of maximum perturbation $\epsilon$ and (3) the per step small perturbation $\alpha$.  
In our experiments we have set $\epsilon=0.1$ heuristically similarly to~\cite{goodfellow2015explaining,kurakin2017adversarial} and the rest of BIM hyperparameters were left at their default value in the Cleverhans API~\cite{papernot2018cleverhans}.

\begin{algorithm}
\begin{algorithmic}[1]
  \caption{Iterative Adversarial Attack}
  \label{algo-bim}
  
 \renewcommand{\algorithmicensure}{\textbf{Parameter:}}
 \renewcommand{\algorithmicrequire}{\textbf{Input:}}
 \renewcommand{\algorithmicreturn}{\textbf{Output:}}
  
  \ENSURE $I$, $\epsilon$, $\alpha$
  \REQUIRE original time series $X$ \& its label $\hat{Y}$
  \RETURN perturbed time series $X^{'}$
  
  \STATE $X^{'}\leftarrow X$\
  \FOR {$i=1$ to $I$}
    \STATE$\eta = \alpha \cdot sign(\nabla_{x}J(X^{'},\hat{Y}))$
    \STATE $X^{'}=X^{'}+\eta$
    \STATE $X^{'}=min\{X+\epsilon,max\{X-\epsilon,X^{'}\}\}$
    \ENDFOR
\end{algorithmic}
\end{algorithm}

\section{Results}

\subsection{Experimental setup}
To train the deep neural network, we leveraged the parallel computation of a cluster of more than 60 GPUs (a mix of GTX 1080 Ti, Tesla K20, K40 and K80). 
All of our experiments were evaluated on the 85 datasets from the publicly available UCR archive~\cite{chen2015ucr}.
The model was trained/tested using the original training/testing splits provided in the archive.  
To perform the attacks, we have adapted the Cleverhans API~\cite{papernot2018cleverhans} by extending the well known attacks for time series data and perturbed only the test instances without using the test labels, similarly to the computer vision literature~\cite{yuan2017adversarial}. 

For reproducibility and to allow the time series community to verify and build on our findings, the source code for generating adversarial time series is publicly available on our GitHub repository\footnote{\url{https://github.com/hfawaz/ijcnn19attacks}}.
In addition, we provide on our companion web page\footnote{\url{https://germain-forestier.info/src/ijcnn2019/}} the raw results, our pre-trained models as well as a set of perturbed time series for each dataset in the UCR archive.
This would allow time series data mining practitioners to test the robustness of their machine learning models against adversarial attacks.

\begin{figure}[t]
\centering
\includegraphics[width=.9\linewidth]{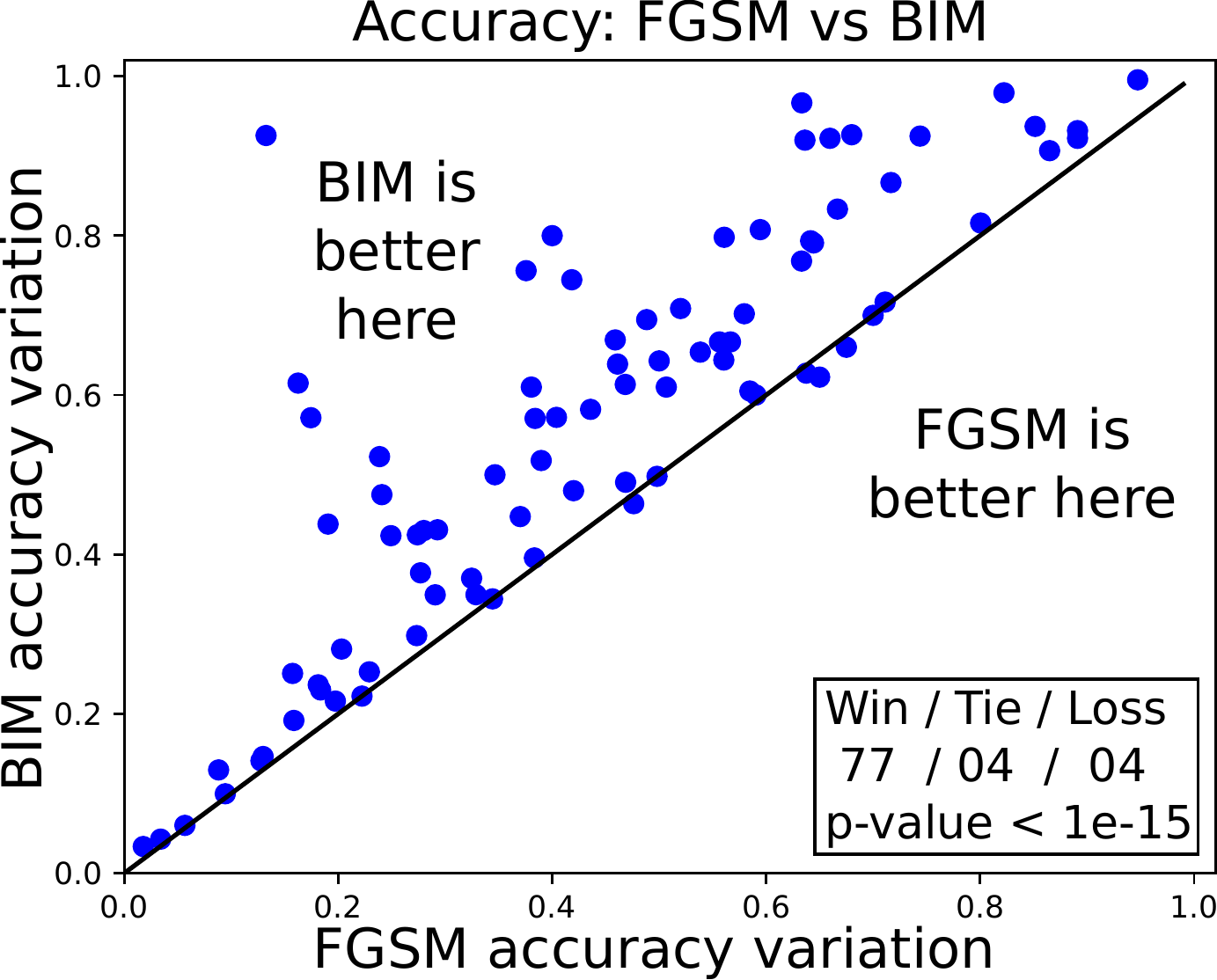}
\caption{Accuracy variation for two attacks (FGSM and BIM) with respect to ResNet's original accuracy.}
\label{fig-acc-plot}
\end{figure}

\subsection{Results on the whole UCR archive}
For all datasets, both attacks managed to reduce ResNet's accuracy. 
One exception is the DiatomSizeReduction dataset which is the smallest one in the archive with an already low original accuracy equal to 30\% due to overfitting~\cite{ismailfawaz2018deep}.
Figure~\ref{fig-acc-plot} shows the accuracy variation for both attacks with respect to the network's original accuracy on the UCR archive.
On average, over the 85 datasets, FGSM and BIM managed to reduce the model's accuracy respectively by 43.2\% and 56.89\%. 
The Wilcoxon signed-rank test indicates that BIM is \emph{significantly} better than FGSM in decreasing the model's accuracy, with a $p$-value~$\le10^{-15}$.  
However, we should note that FGSM is a fast approach allowing real-time generation of adversarial time series whereas BIM is time-consuming and requires a certain number of iterations $I$. 

By analyzing the use-cases where both attacks failed to fool the classifier, we found out that the corresponding datasets have two interesting characteristics that could explain the classifier's robustness to adversarial examples. 
The first one is that 50\% of the simulated datasets (CBF, Two\_patterns and synthetic\_control) in the archive are in the top six hardest datasets to attack.
Perhaps since these are synthetic datasets generated by humans to serve some human intuition for TSC, small perturbations imperceptible by humans, are not enough to alter the classifier's decision.
The second observation is that a network trained on datasets with time series of short length is harder to fool. 
This is rather expected since the less data points we have, the less amount of perturbation the attacker is allowed to add. 
For example ItalyPowerDemand contains the shortest sequences ($T=24$) and is the second most hardest use-case for both attacks.

\subsection{Multi-Dimensional Scaling}

We used Multi-Dimensional Scaling (MDS)~\cite{kruskal1978multidimensional,forestier2017generating} with the objective to gain some insights on the spatial distribution of the perturbed time series compared to the original ones. 
MDS uses a pairwise distance matrix as input and aims at placing each object in an N-dimensional space such as the between-object distances are preserved as well as possible.
Using the Euclidean Distance (ED) on a set of time series (original and perturbed), it is then possible to create a similarity matrix and apply MDS to display the set into a two dimensional space.
The latter straightforward approach supposes that the ED is able to strongly separate the raw time series, which is usually not the case evident by the low accuracy of the nearest neighbor when coupled with the ED~\cite{bagnall2017the}. 
Therefore, we decided to use the \emph{linearly} separable representation of time series from the output of the GAP layer, which is used as input to the softmax \emph{linear} classifier (multinomial logistic regression). 
More precisely, for each input time series, the last convolution outputs a multivariate time series whose dimensions are equal to the number of filters (128) in the last convolution, then the GAP layer averages the latter 128-dimensional multivariate time series over the time dimension resulting in a vector of 128 real values over which the ED is computed. 
This enables the MDS projection to be as close as possible to ResNet's latent representation of the time series. 
As we worked with the ED, we used Metric MDS~\cite{kruskal1978multidimensional} that minimizes a cost function called \emph{Stress} which is a residual sum of squares:

\begin{equation}
    Stress_D(X_1,\ldots,X_N)=\Biggl(\frac{\sum_{i,j}\bigl(d_{ij}-\|x_i-x_j\|\bigr)^2}{\sum_{i,j}d_{ij}^2}\Biggr)^{1/2}
\end{equation}
where $d_{ij}$ is the ED between the GAP vectors of time series $X_i$ and $X_j$.
Obviously, one has to be careful about the interpretation of MDS output, as the data space is highly simplified (each time series $X_i$ is represented as a single data point $x_i$).

\subsection{Attacks on food quality and safety}

\begin{figure}[t]
\centering
\includegraphics[width=.9\linewidth]{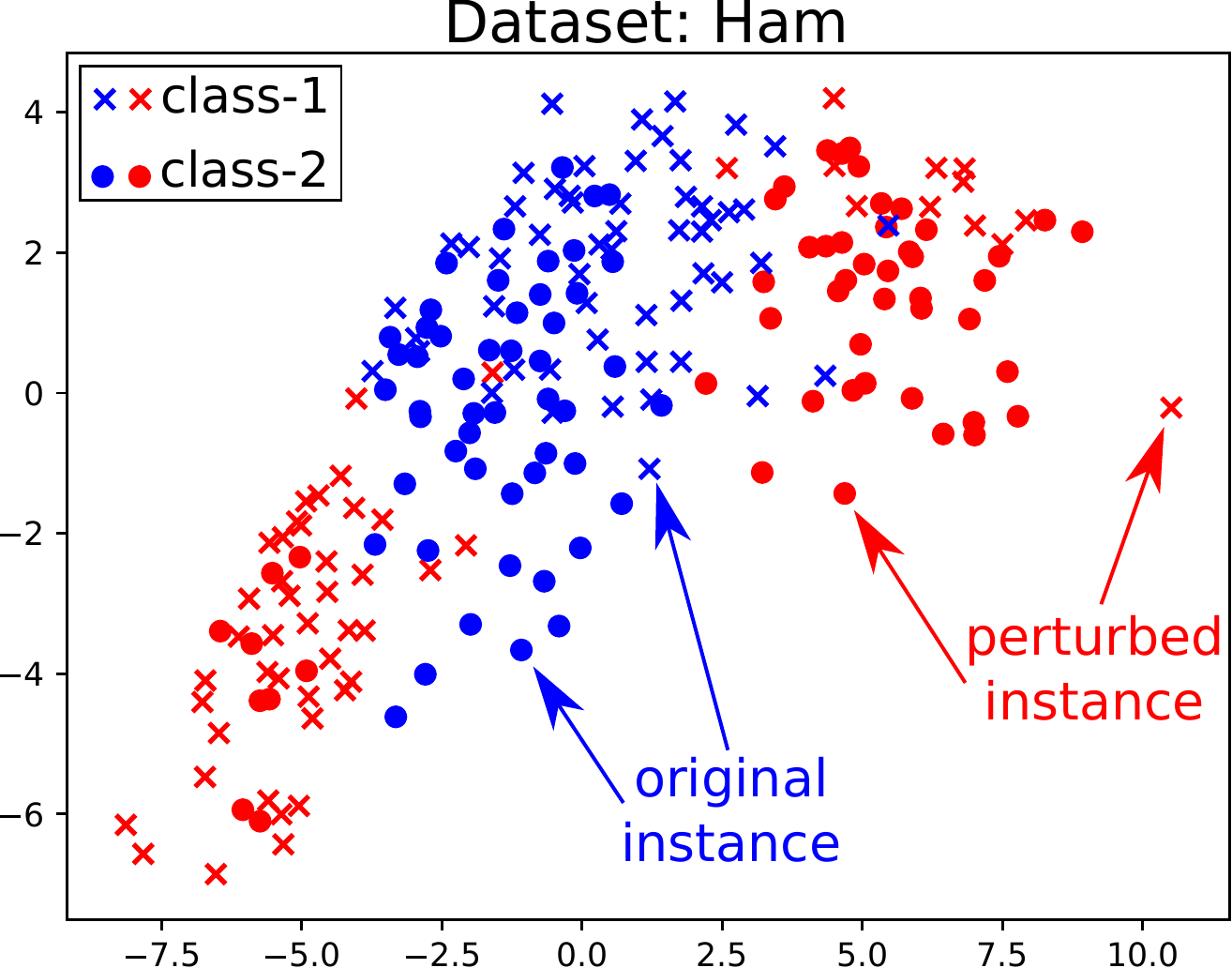}
\caption{Multidimensional Scaling (MDS) showing the distribution of perturbed time series on the whole test set of the Ham dataset where the accuracy decreased from 80\% to 21\% after performing the BIM attack.}
\label{fig-ham-mds}
\end{figure}

The determination of food quality, type and authenticity along with the detection of adulteration are major issues in the food industry~\cite{nawrocka2013determination}.
With meat related product, authenticity checking concerns for example the identification of substitution of high value raw materials with cheaper materials like less costly cuts, offal, blood, water, eggs or other types of proteins.
These substitutions not only decease the consumers but can also cause severe allergic responses as the substitute materials are hidden.
Discriminating meat that has been frozen-and-thawed from fresh meat is also an important issue, as refreezing food can result in an increased amount of bacteria.
Spectroscopic methods have been historically very successful at evaluating the quality of agricultural products, especially food~\cite{nawrocka2013determination}.
This technique is routinely used as a quality assurance tool to determine the composition of food ingredients.

In this context, an adversarial attack could be used to modify recorded spectrographs (seen as time series) in order to hide the low qualify of the food.
The Beef dataset~\cite{al2002detection} (from the UCR archive) contains four classes of beef spectrograms, from pure and adulterated beef with varying degrees of potential adulterants (heart, tripe, kidney, and liver).
An adversarial attack could thus consist in modifying an adulterated beef to make a network classify it as pure beef.
For this dataset, FGSM and BIM reduced the model's accuracy respectively by 56.7\% and 66.7\%. 


The Ham dataset~\cite{olias2006sodium} contains measurements from 19 Spanish and 18 French dry-cured hams, with the goal to distinguish the provenance of the food.
An adversarial attack could consist in perturbing the spectrograms to hide the real provenance of the food. 
Figure~\ref{fig-ham-mds} shows the MDS projection of the original and perturbed instances for Ham's test set, where one can see that the adversarial examples are \emph{pushed} toward the other class.

\begin{figure}[t]
\centering
\includegraphics[width=0.9\linewidth]{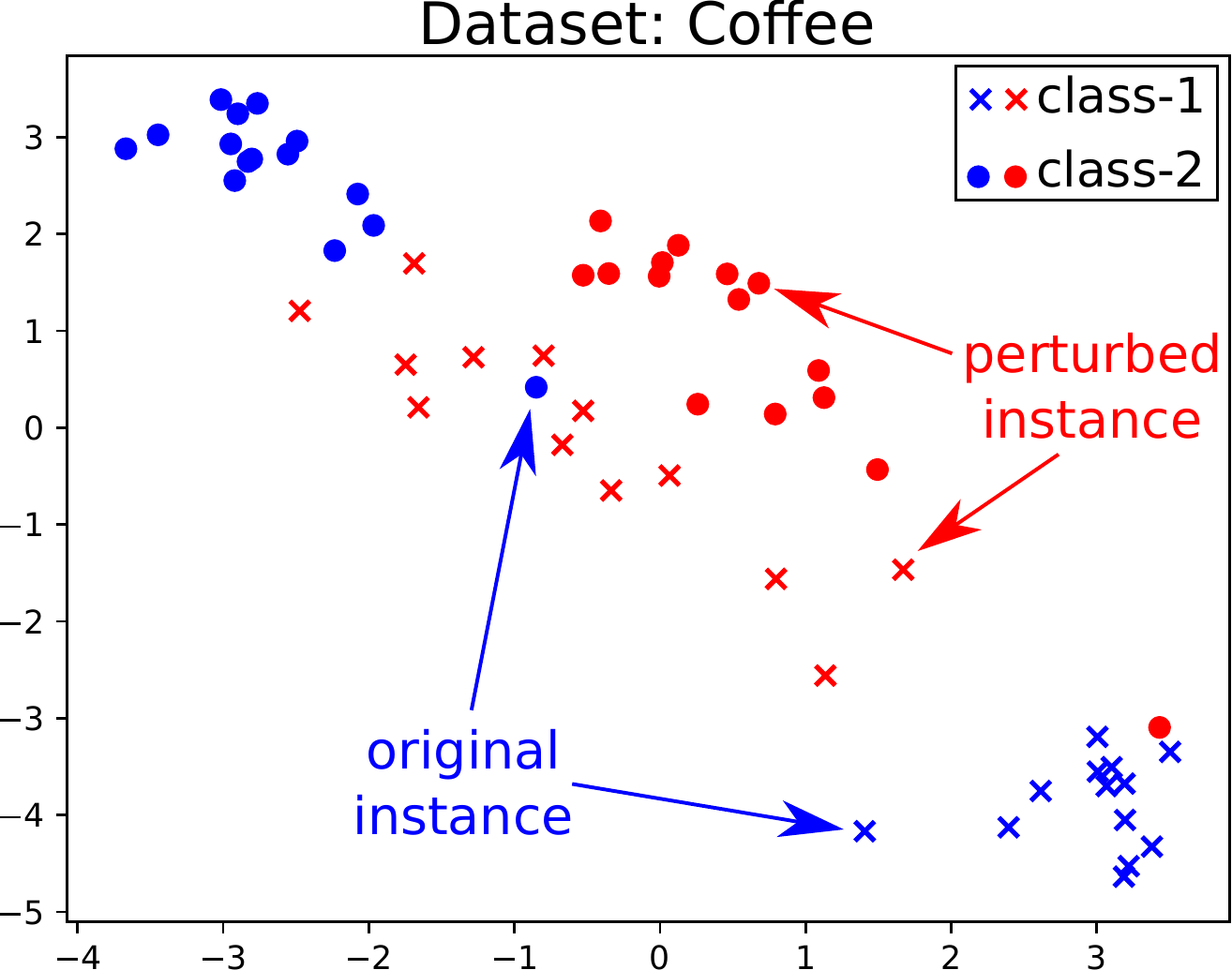}
\caption{Multidimensional Scaling (MDS) showing the distribution of perturbed time series on the whole test set of the Coffee dataset where the accuracy decreased from 100\% to 50\% after performing the FGSM attack.}
\label{fig-coffee-mds}
\end{figure}

The Coffee dataset~\cite{briandet1996discrimination} is a two class problem to distinguish between Robusta and Arabica coffee beans.
Arabica beans are valued most highly by the trade, as they are considered to have a finer flavor than Robusta.
An adversarial attack could consist in altering the spectrograms to make Robusta beans look like Arabica  beans.
Figure~\ref{fig-coffee-mds} shows the MDS representation of the original and perturbed time series from the test set. 
We can clearly see how the instances are \emph{pushed} toward the class frontiers after performing the FGSM attack. 

\subsection{Attacks on vehicle sensors}

The increase in the number of sensors and other electrical devices has drastically augmented the amount of data produced in the industry.
These data are now routinely used to monitor systems or to perform predictive maintenance and prevent failures~\cite{susto2015machine}.
The car industry is not an exception with the increasing number of sensors present in modern vehicles, especially for advanced driver assistance systems and autonomous driving.

Data are also used to perform diagnostic on vehicles in order to detect engine problems or compliance with environmental regulations.
In this context, an adversarial attack could consist in altering sensor readings in order to hide a specific problem or to pass a CO$_2$ emission test.
The famous ``dieselgate'' (or ``emissionsgate'')~\cite{brand2016beyond} made this kind of attack a reality as multiple automakers have been suspected of using emission control systems during laboratory emissions testing.

To illustrate this use case, we used the FordA datasets (from the UCR archive) that was was originally used in a competition in the IEEE World Congress on Computational Intelligence, 2008.
The classification problem is to diagnose whether a certain symptom exists or not in an automotive subsystem.
Each case consists of 500 measurements of engine noise and a class label.
In this context, an attack could consist in hiding an engine problem.
In practice for the FordA dataset, the model's accuracy decreased by 57.9\% and 70.2\% when applying respectively the FGSM and BIM attacks.  

\begin{figure}[t]
\centering
\includegraphics[width=0.9\linewidth]{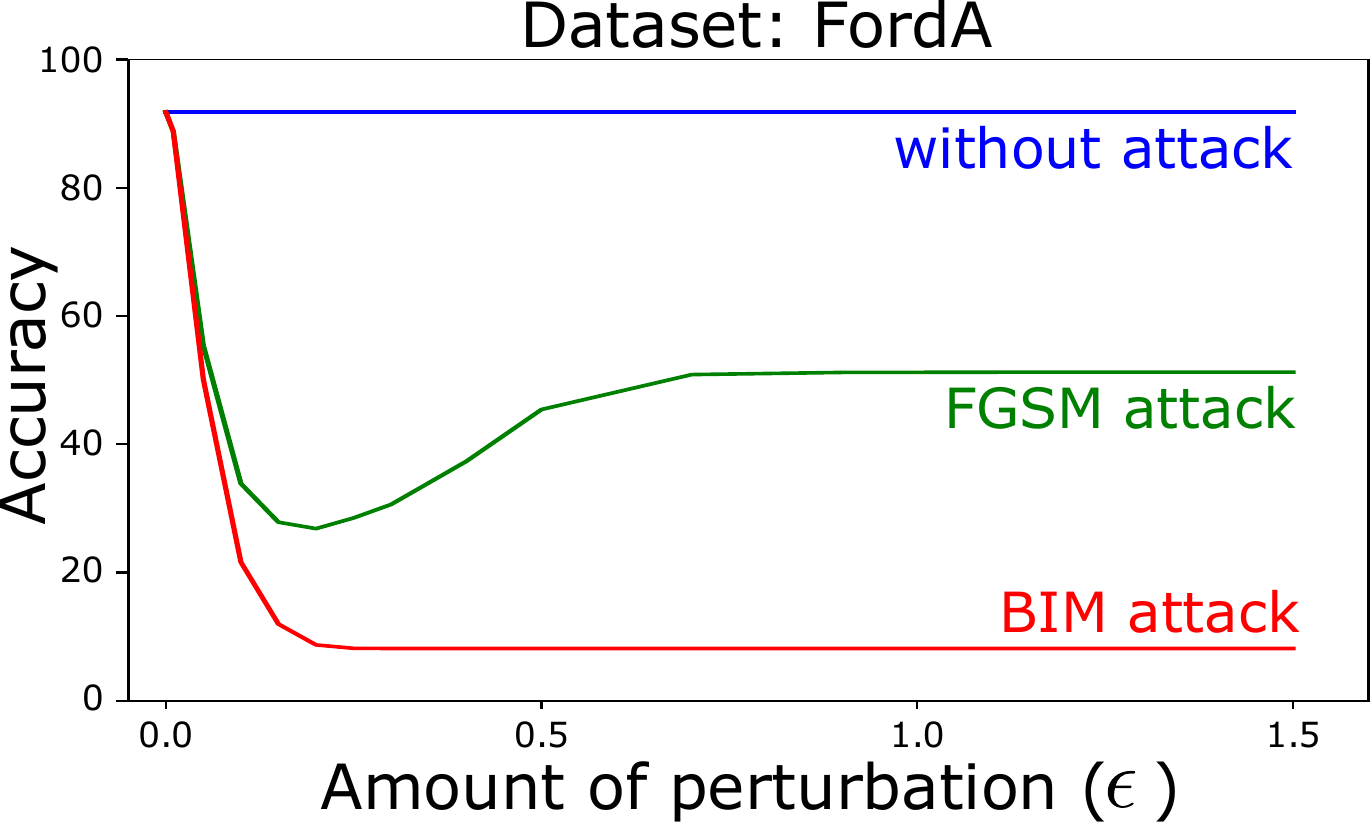}
\caption{Accuracy variation with respect to the amount of perturbation for FGSM and BIM attacks on FordA.}
\label{fig-plot-eps}
\end{figure}

Figure~\ref{fig-plot-eps} depicts the variations of ResNet's accuracy on FordA with respect to the amount of perturbation $\epsilon$ allowed for the FGSM and BIM attacks. 
As expected~\cite{kurakin2017adversarial}, we found that FGSM fails to generate adversarial examples that can fool the network for larger values of $\epsilon$, whereas the BIM produces perturbed time series that can reduce a model's accuracy to almost 0.0\%.
This can be explained by the fact that BIM adds a small amount of perturbation $\alpha$ on each iteration whereas FGSM adds $\epsilon$ amount of noise for each data point in the series that may not be useful for misclassifying the test sample. 

\subsection{Attacks on electricity consumption}
 
Smart meters are electronic devices that record electric power consumption while sending information to the electricity supplier for monitoring, billing and data analysis. 
These meters typically register energy hourly and report back at least once a day to the supplier by leveraging a two-way communication channel between the device and the supplier's central system. 
These smart meters have raised a set of concerns in public opinion especially because they send detailed information about how much electricity is being used for each time stamp. 
Precisely, it has been shown that it is possible to know exactly which type of electric device is or has been used from simply analyzing the electricity consumption data~\cite{owen2012powering}.
In this context, an attack could consist in modifying the electricity consumption time series of one device to make it recognized as another in order to hide which devices are actually used by a specific user.

To illustrate this use case, we used the SmallKitchenAppliances dataset from the UCR archive that was recorded as part of government sponsored study called Powering the Nation~\cite{owen2012powering}.
By collecting and analyzing behavioural data about consumers' daily use of electricity within their homes, the goal is to reduce the UK's carbon footprint. 
The dataset contains readings from 251 households recorded over a month. 
Each univariate time series has a length equal to 720 corresponding to 24 hours of readings taken every two minutes. 
The three classes are: Kettle, Microwave and Toaster.
For this dataset, FGSM and BIM managed to reduce the classifier's accuracy respectively by 38.4\% and 57\%. 

\begin{figure}[t]
\centering
\includegraphics[width=.98\linewidth]{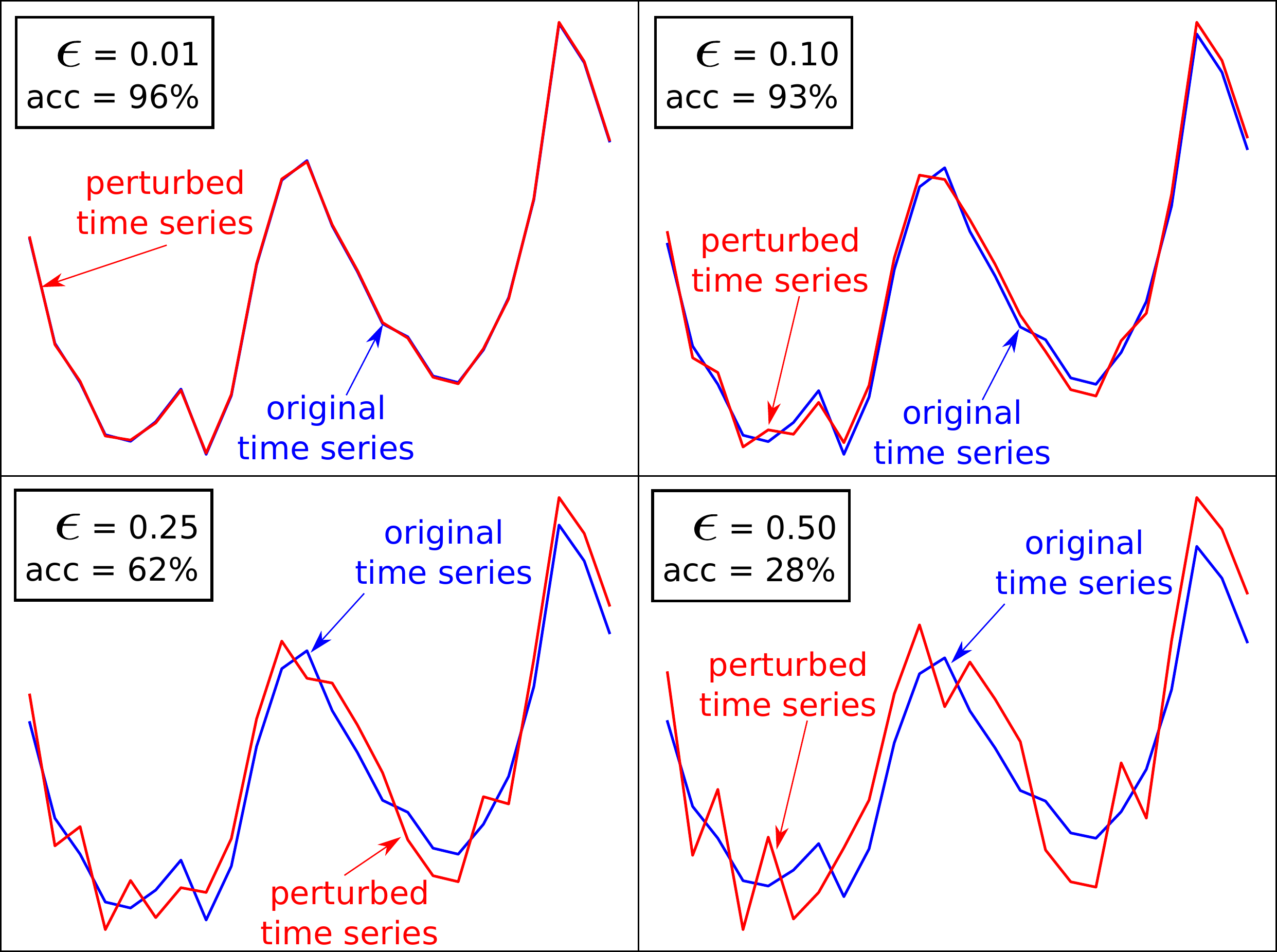}
\caption{
Accuracy variation for ItalyPowerDemand with respect to the perturbation $\epsilon$ where FGSM managed to fool the network with this example for $\epsilon\ge 0.3$. 
}
\label{fig-pert-epsilon}
\end{figure}

The ItalyPowerDemand dataset~\cite{keogh2006intelligent}, another dataset from the UCR archive, contains twelve monthly electrical power demand time series from Italy. 
The task is to differentiate between instances that correspond to winter months (October to March) and summer months (April to September). 
This dataset contains the shortest time series in the archive ($T=24$), thus requiring a higher amount of perturbation $\epsilon$ in order to be misclassified. 
Figure~\ref{fig-pert-epsilon} shows the variation of the model's accuracy as well as the shape of a time series from the ItalyPowerDemand dataset with respect to the amount of noise that is added. 
For this example, both attacks needed higher values of perturbation ($\epsilon\ge0.3$) rather than the default setting ($\epsilon=0.1$). 

\subsection{Transferability of adversarial examples} 
To evaluate the transferability of perturbed time series, we used the Fully Convolutional Neural Network (FCN) which was originally proposed in~\cite{wang2017time} and was recently shown to be the second most accurate deep time series classifier~\cite{ismailfawaz2018deep} when evaluated on the UCR archive~\cite{chen2015ucr}.
We used the test sets altered with FGSM and BIM using ResNet and try to classify it with FCN (both were originally trained on the same train set).
For both FGSM and BIM attacks, FCN's accuracy decreases respectively by 38.2\% and 42.8\% which shows that adversarial examples are capable of generalizing to a different network architecture. 
The Wilcoxon signed-rank test also shows that BIM is \emph{significantly} better than FGSM in reducing FCN's accuracy, with a $p$-value $\le10^{-7}$.
This type of attacks is known as ``black box'' where the attackers do not have access to the target model's internal parameters (FCN) yet they are able to generate perturbed time series that fool the classifier.

\subsection{How can we prevent such attacks ?}
Countermeasures for adversarial attacks~\cite{yuan2017adversarial} follow two defense strategies: (1) \emph{reactive}: identify the perturbed instance; (2) \emph{proactive}: improve the network's robustness without generating adversarial examples. 
One of the most straightforward proactive methods is \emph{adversarial training}, which consists of (re)training the classifier with adversarial examples. 
Other reactive techniques consist of detecting the adversarial examples during testing.
However, most of these detectors are still prone to attacks that are designed specifically to fool the detectors~\cite{yuan2017adversarial}. 
Therefore, we think that the time series community would have much to offer in this area by leveraging the decades of research into non-probabilistic classifiers such as the nearest neighbor coupled with DTW~\cite{tan2018efficient}.  
Running classifiers against the adversarial examples that we provide in this paper is a first step toward identifying vulnerable models and making them more robust to such type of attacks.



\section{Conclusion}
In this paper, we introduced the concept of adversarial attacks on deep learning models for time series classification.
We defined and adapted two attacks, originally proposed for image recognition, for the TSC task.   
We showed how adversarial perturbations are able to reduce the accuracy for the state-of-the-art deep learning classifier (ResNet) when evaluated on the UCR archive benchmark. 
With deep neural networks becoming frequently adopted by time series data mining practitioners in real-life critical decision making systems, we shed the light on some crucial use cases where adversarial attacks could have serious and dangerous consequences. 

In the future, we would like to investigate countermeasures techniques to defend machine learning models against such attacks while exploring the transferability of adversarial examples to other non deep learning state-of-the-art classifiers.
Finally, we would like to further explore the dozens of adversarial attacks that are published each year in order to identify and protect vulnerable deep learning models for TSC. 

\section*{Acknowledgements}
We would like to thank the providers of the UEA/UCR datasets, as well as NVIDIA Corporation for the Quadro P6000 grant and the M\'esocentre of Strasbourg for providing access to the cluster.

\bibliographystyle{IEEEtran}
\bibliography{biblio}

\end{document}